\documentclass[journal]{IEEEtran}
\usepackage{latexsym}
\usepackage{graphicx}
\usepackage{amsfonts,amssymb,amsmath}
\usepackage{hyperref}
\usepackage{cite}
\usepackage{textcomp}
\usepackage{amsthm}
\usepackage{booktabs}
\usepackage{algorithm}
\usepackage{algorithmic}
\usepackage{subfig}
\usepackage{multirow}

\begin{document}

\title{Federated Self-Supervised Learning of Multi-Sensor Representations for Embedded Intelligence}
\author{Aaqib Saeed, Flora D. Salim~\IEEEmembership{Member,~IEEE}, Tanir Ozcelebi~\IEEEmembership{Member,~IEEE}, and Johan Lukkien~\IEEEmembership{Senior Member,~IEEE}
\thanks{Aaqib Saeed, Tanir Ozcelebi and Johan Lukkien are with the Department of Mathematics and Computer Science, 
Eindhoven University of Technology, The Netherlands. E-mail: \{a.saeed, t.ozcelebi, j.j.lukkien\}@tue.nl. Correspondence should be addressed to Aaqib Saeed.}
\thanks{Flora D. Salim is with the School of Science, RMIT University, Melbourne Australia. She co-directs the RMIT Centre for Information Discovery and Data Analytics (CIDDA). E-mail: flora.salim@rmit.edu.au.}
\thanks{This work is funded by SCOTT (www.scott-project.eu) project. It has received funding from the Electronic Component Systems for 
European Leadership Joint Undertaking under grant agreement No 737422. This Joint Undertaking receives support from the European Union's 
Horizon 2020 research and innovation programme and Austria, Spain, Finland, Ireland, Sweden, Germany, Poland, Portugal, Netherlands, Belgium, Norway.}
}

\markboth{IEEE INTERNET OF THINGS JOURNAL}{}
\maketitle

\begin{abstract}
Smartphones, wearables, and Internet of Things (IoT) devices produce a wealth of data that cannot be accumulated in a centralized repository for learning supervised models due to privacy, bandwidth limitations, and the prohibitive cost of annotations. Federated learning provides a compelling framework for learning models from decentralized data, but conventionally, it assumes the availability of labeled samples, whereas on-device data are generally either unlabeled or cannot be annotated readily through user interaction. To address these issues, we propose a self-supervised approach termed \textit{scalogram-signal correspondence learning} based on wavelet transform to learn useful representations from unlabeled sensor inputs, such as electroencephalography, blood volume pulse, accelerometer, and WiFi channel state information. Our auxiliary task requires a deep temporal neural network to determine if a given pair of a signal and its complementary viewpoint (i.e., a scalogram generated with a wavelet transform) align with each other or not through optimizing a contrastive objective. We extensively assess the quality of learned features with our multi-view strategy on diverse public datasets, achieving strong performance in all domains. We demonstrate the effectiveness of representations learned from an unlabeled input collection on downstream tasks with training a linear classifier over pretrained network, usefulness in low-data regime, transfer learning, and cross-validation. Our methodology achieves competitive performance with fully-supervised networks, and it outperforms pre-training with autoencoders in both central and federated contexts. Notably, it improves the generalization in a semi-supervised setting as it reduces the volume of labeled data required through leveraging self-supervised learning. 
\end{abstract}


\section{INTRODUCTION}
\label{sec:introduction}
\IEEEPARstart{L}{earning} representations with deep neural networks have made tremendous improvements in the last few years on challenging real-world tasks~\cite{oord2018representation,owens2018audio,mikolov2013distributed, radu2018multimodal}, thanks to the emergence of massive datasets. In particular, the wealth of sensory data from the Internet of Things (IoT) devices are only recently being leveraged for tackling important problems in understanding context, user monitoring, health, and other predictive analytics tasks, e.g., for emotional well-being~\cite{schmidt2018introducing, ballinger2018deepheart}, sleep tracking~\cite{supratak2017deepsleepnet}, and physical activity detection~\cite{saeed2019multi}. The success is mainly attributed to the supervised methods that utilize labeled datasets for training models in a central environment. In contrast, learning models from unlabeled decentralized data still presents a major challenge. Obtaining large, well-curated sensory data from edge devices is especially difficult owing to issues like user privacy, the prohibitive cost of labeling, bandwidth limitations, network connectivity, and the diversity of device types~\cite{lane2016deepx}. These factors make it significantly challenging to harness abundant data on remote devices for learning semantic features with standard supervised approaches. 

To highlight the challenges associated with learning a generalizable model for a particular use case, consider this illustrative example. Let us assume that we aim to develop a robust sleep stage classification model that can be used for a larger population of users. The standard methodology is to learn a supervised model and requires example-label pairs for providing supervision so that a model can differentiate among instances of multiple classes through learning underlying patterns in the input. The procedure begins with the data collection to monitor hundreds of users for electroencephalography (EEG) or other signals as they progress through various stages of sleep and accumulate the multi-sensor data in a centralized (data center) repository for further analysis. The next step is then to get the aggregated inputs annotated by the sleep expert (i.e., generally a professional trained in analyzing physiological signals) for the sleep classes, such as wake, N$1$, N$2$, N$3$, and rapid eye movement. Then, the learning and evaluation phase involves several iterations of improving the model performance. Lastly, the model is deployed in the wild for user monitoring. 
The process of model development, from data collection to annotation, could be extremely costly owing to the difficulty in setting up an experimental (data collection) protocol. Furthermore, the domain expertise required for the labeling could be severely limited. This problem is exacerbated by the need of supervised deep neural network models for a massive amount of labeled data to learn discriminative features. It becomes painstakingly difficult to inspect and annotate hundreds of thousands of hours of multi-sensor data. Therefore, in practice, limited-sized sensor data are collected and labeled for learning the model, which could further affect its generalization. The important point to note here is that the explained strategy is only applicable when the users agree on sharing their data for learning, which is not ideal in several real-world contexts due to raising privacy issues and misaligned incentives for the user. Likewise, IoT devices produce an astonishing amount of data on a daily basis, and even if the data sharing takes place, its rapidly increasing size limits exploiting for learning models. Therefore, there is a need to develop unsupervised (or self-supervised) methods that can be used to learn general-purpose models from unlabeled data. It is particularly pertinent to on-device learning (such as a smartphone), without the need for data aggregation in a centralized server, and minimal to no human involvement in terms of the annotation process. Consequently, the unsupervised model can be used as a semantic feature extractor or initialization for efficiently adapting to an end-task of interest through fine-tuning with few-labeled instances.   

Specifically, the aforementioned challenges motivate the following research questions: Can we train a deep network to extract useful sensory representations in an unsupervised manner without utilizing strong labels for a specific problem, such as activity recognition? Could it also be achieved without aggregating the local data samples from remote devices in a centralized repository, i.e., employing decentralized or on-device learning? Can we improve the network generalization in a low-data regime through fine-tuning it with few-labeled examples that potentially could be easily pooled from a group of users?

Previous approaches to learning representations from time-series of sensory modalities with deep networks can be mainly categorized into three areas: end-to-end training of supervised models with labeled data~\cite{supratak2017deepsleepnet, yousefi2017survey, ballinger2018deepheart, radu2018multimodal}, reconstruction of the actual input for pre-training~\cite{vincent2008extracting, martinez2013learning, wang2019deep}, and utilizing self-learning with domain-specific transformations or cross-modal learning~\cite{baltruvsaitis2018multimodal}. Primarily, the focus of these methods is to conduct training of predictive models on a central server in a data center. However, as mentioned earlier, the aggregation of continuously increasing data from distributed devices is practically infeasible, aside from privacy issues. Initially, geo-distributed analytics~\cite{vulimiri2015wanalytics, dean2008mapreduce} and distributed learning~\cite{zhang2015deep, balcan2012distributed, shamir2014distributed} in a data center environment is studied to exploit data locality and reducing network costs through pushing code to the data on the edge which generally is a node in the data center which could be across the globe. Nevertheless, these methods do not address the fundamental problem of learning representations with deep networks from unlabeled and highly distributed data that resides on user devices, which can not be aggregated in a central environment for learning. 

To address the aforementioned concerns, federated learning~\cite{44822} is emerging as an effective way of collaboratively training shared models from distributed private data. However, existing exploration in this area is solely focused on learning supervised models for tasks where annotations can be easily acquired based on the user interaction, e.g., mobile keyword prediction~\cite{hard2018federated}. The curation of strongly labeled data becomes infeasible as annotations can not be acquired easily for solving several important problems involving sensory inputs. Because apart from wearables, other sensors could be installed in remote locations, and expert-level domain knowledge could be required to annotate samples. Hence, in such cases, unsupervised approaches provide a compelling substitute to learn from unlabeled data available in huge quantities as they do not require semantic labels.  

One of the most rudimentary forms of unsupervised feature discovery has been hand-crafted feature engineering, which turns out to be largely redundant due to its limited discriminative power for building high-performance models~\cite{lecun2015deep}. Another area of research that is considerably explored focuses on reconstruction based approaches for extracting low-dimensional embedding through learning from data with deep autoencoders~\cite{vincent2008extracting}. The main drawback of these methods is that they may waste the network's capacity to model low-level input details through predicting every bit of the signal. This is not needed if the aim is to learn discriminatory features that generalize well to the downstream (or end) tasks, e.g., sleep stage classification with electrical brain activity signals. 

A promising substitute is the emerging area of self-supervised learning~\cite{de1994learning}, which enables the learning of representations through solving an auxiliary task for which labels can be acquired from the data without any human intervention. In this case, several techniques are proposed mainly for audio, visual and textual data including estimation of missing input~\cite{zhang2017split}, prediction of contextually relevant information~\cite{noroozi2016unsupervised}, recognizing degree of rotation applied on an image~\cite{gidaris2018unsupervised}, contrastive predictive coding~\cite{oord2018representation}, synchronization of audio-visual inputs~\cite{owens2018audio}, and robotic imitation using multi-view videos~\cite{sermanet2018time}. Moreover, cross-modal learning is also utilized by specifying an appropriate loss term between different input modalities for training multimodal networks. However, to the best of our knowledge, previous work did not study self-supervised learning for other sensing modalities (e.g., electroencephalography, accelerometer, blood volume pulse, and others) as produced by a variety of IoT devices. 

\begin{figure}[t]
\centering
\subfloat{\includegraphics[width=6.5cm]{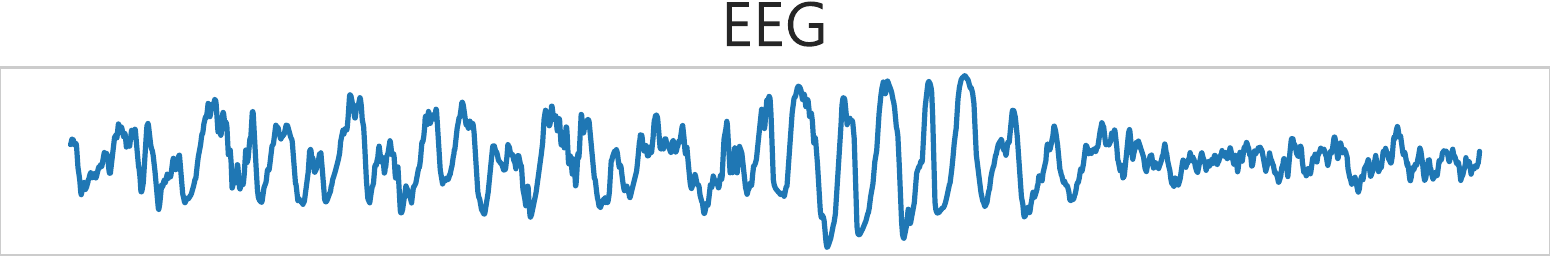}} \\
\subfloat{\includegraphics[width=6.5cm]{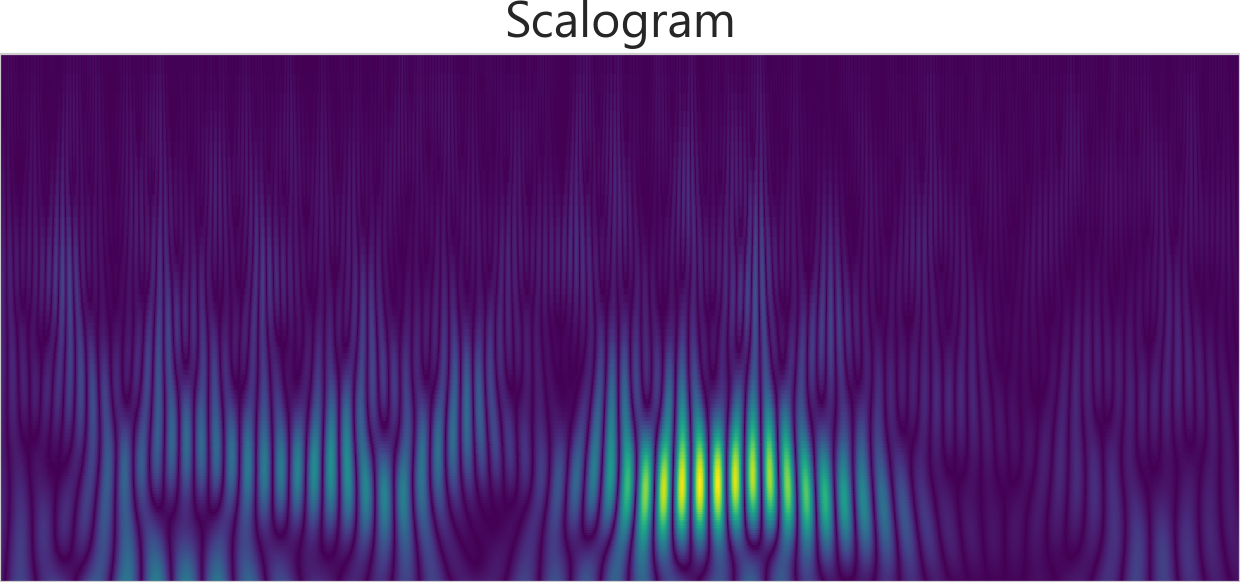}}
\caption{\small{Illustration of a $30$-seconds long electroencephalogram (EEG) signal and a corresponding scalogram extracted with a Morlet wavelet transform.}}
\label{fig:signal_scalogram}
\end{figure}

In this work, we hypothesize that the fusion of self-supervision with federated learning could result in an effective method for learning from unlabeled, private, and diverse types of sensory data, which is crucial for several embedded (personalized) machine learning tasks. To achieve this objective, we develop a novel auxiliary task based on a wavelet transform, which we call \textit{scalogram-signal correspondence learning} (SSCL). A deep temporal convolution network is trained to solve the specified task so as to learn representations from a variety of sensory inputs (e.g., electroencephalography, inertial measurement unit's sensors (IMUs), and WiFi channel state information). We name it a \textit{scalogram contrastive network} (SCN). Specifically, the self-supervised scheme is designed to contrast between a raw signal (time-series) and its complementary view, which, in our case, is a scalogram, extracted with continuous wavelet transform~\cite{merry2005wavelet}. We note that other views, such as a spectrogram derived with a fast Fourier transform can also be used for this purpose (or in combination). In this work, we opt for wavelet transformation because it is better at localizing time-frequency properties~\cite{daubechies1990wavelet} of the signal.

The core idea behind our pretext task is to determine if a given pair of scalogram-signal inputs are aligned or misaligned, i.e., whether a scalogram is the transformation of a given signal. The presented auxiliary task can formally be seen as a binary classification problem, and we employ a contrastive objective inspired by~\cite{chopra2005learning} for optimizing it (see Figure~\ref{fig:wcn} for an overview) in both central and federated settings without involving a human in the data labeling process. Importantly, we would like to highlight that for the model to solve the defined task successfully, it should learn the core semantics in shared input views through possibly relating frequency, scale, and other information present in the signal. The network captures meaningful latent relationships through correlating scalogram-signal inputs in the embedding space. Mainly, the representations that could emerge from the learning process are forms of invariances (such as sensor noise, subject-specific variations), which are essential in several tasks involving sensory data, e.g., stress detection with physiological signals. 

The key contributions of this work are three-fold: First, we propose a scalogram-signal correspondence learning framework for self-supervised learning from diverse sensory data. Second, to the best of our knowledge, we, for the first time, propose to unify federated learning with self-supervision to learn from unlabeled and private data on edge devices. Third, we extensively assess the proposed method on several publicly available datasets from different domains with linear classification protocol in central and federated contexts, low-data regime (i.e., semi-supervised setting), and transfer learning including cross-validation. The SCN achieves competitive performance compared with fully-supervised networks that are trained entirely on labeled data and perform significantly better than other approaches. Particularly, SCN fine-tuning with few-labeled instances, e.g., five or ten instances per class, improves the F-score by as much as $5$\%-$6$\% in comparison to training from scratch. Our approach also works better than transferring supervised features, learned from the source data, between the related tasks.

\section{BACKGROUND and RELATED WORK}
\label{sec:bgrw}
We consider learning sensory features from raw unlabeled data with a deep neural network $\mathcal{F}_{\theta}$ (parameterized by $\theta$), which transforms input from $\mathcal{X}$ into output in $\mathcal{Z}$. Here, we refer to a vector obtained through applying a mapping function $\mathcal{F}: \mathcal{X} \mapsto \mathcal{Z}$ from an arbitrary intermediate or penultimate layer of the network as `representation' or `feature.' Our objective is to learn general-purpose representations that can make subsequent tasks of interests easier to solve. To this end, numerous unsupervised methods are developed to leverage a large amount of unlabeled data for achieving better generalization. Moreover, the data required for model development could not only be unannotated but also distributed, without the option to accumulate it in a centralized repository due to privacy concerns and its ever-increasing size. To tackle the issue of learning models from decentralized user data, the field of federated learning~\cite{44822} is rapidly gaining momentum. Our work is intended to unify self-supervision with federated learning to realize the vision of on-device learning, with a focus on multi-sensor inputs. We describe the details of the essential building blocks of our approach and related work in the following subsections.

\subsection{Self-supervised Learning}
The field of unsupervised learning deals with extracting disentangled representations that could be used for solving a wide variety of end-tasks. The most prominent approaches include principal component analysis, Boltzmann machine~\cite{salakhutdinov2009deep}, autoencoders~\cite{vincent2008extracting}, generative adversarial networks~\cite{goodfellow2014generative}, and autoregressive models~\cite{oord2016wavenet}. Another emerging area of research for extracting unsupervised representations is `self-supervision.' It provides a general and powerful framework for learning with unlabeled inputs through solving pretext tasks. Here, a surrogate objective is specified in such a way that optimizing it would force the network to learn meaningful and usable features for the end-task. Specifically, given an unlabeled dataset $\mathcal{D} = \{x_1, x_2, \ldots, x_\mathcal{M}\}$ with $\mathcal{M}$ instances. A surrogate task is designed that provides pseudo-labels $\{y_1, y_2, \ldots, y_\mathcal{M}\}$ to learn $\mathcal{F}_{\theta}$ (without the need of any strong class annotations) through minimizing a classification, regression or metric loss $\mathcal{L}$ given by:
\begin{align}
   \min_{\theta} \frac{1}{\mathcal{M}} \sum^{\mathcal{M}}_{m=1} \mathcal{L}(\mathcal{F}_{\theta}(x_m), y_m)
\end{align}

In the past few years, several self-supervised methods have been developed for vision, audio, language modeling, and other domains. However, little to no attention is paid towards exploring other sensing modalities, such as electroencephalography, IMUs, and blood volume pulse. The prominent approaches for learning from traditional input modalities include, colorization of grayscale images~\cite{larsson2017colorization}, predicting relative location of an image patch~\cite{noroozi2016unsupervised}, audio-visual synchronization~\cite{owens2018audio}, temporal alignment in videos through cycle-consistency~\cite{dwibedi2019temporal}, word2vec (and other variants)~\cite{mikolov2013distributed}, signal transformation prediction~\cite{saeed2019multi}, contrastive predictive coding~\cite{oord2018representation}, and robotic imitation learning via time-contrastive networks~\cite{sermanet2018time}. These are some of the many strategies proposed for learning from an unlimited amount of unlabeled audio, visual, and textual data. 

In this work, we seek to learn representations from data produced by sensors (time-series) on edge as obtaining a large amount of such labeled data is time-consuming and extremely costly. To solve this problem, we utilize a contrastive objective between a raw and complementary view of the data acquired via wavelet transform. A detailed explanation of the approach is provided in Section~\ref{sec:approach}. 

\subsection{Wavelet Transform}
While the Fourier Transform (FT) sheds light on the frequency properties of the transformed signal, the input signal's time properties are not directly accessible from the Fourier representation. An alternative to this, which provides information about the time properties of the input signal (time locality of signal variations), is the Wavelet Transform (WT)~\cite{daubechies1990wavelet}. Like the Short-term Fourier Transform (STFT), the WT divides the input signal into time windows of a certain size and operates on each time window separately. Choosing a larger time window of WT gives better frequency resolution of the WT output signal, while this reduces the time resolution. Precisely, the Wavelet series gives individual coefficients of a set of orthonormal functions (wavelets, e.g., Morlet, Haar, Daubechies). Like its counterparts, this representation effectively decomposes the input signal into combinations of wavelets. Due to these compelling properties, WT has been widely used in a myriad of domains~\cite{merry2005wavelet}. In particular, continuous WT gained significant popularity compared to discrete counterpart since it is better at localizing time-frequency properties. A wavelet transform of a signal $x(t)$ is defined as follows:
\begin{align}
   \label{eq:wt}
   T(a,b) = \frac{1}{\sqrt{a}} \int_{-\infty}^{+\infty} x(t) \cdot \psi \left( \frac{t - b}{a} \right) dt
\end{align}

\noindent where $\psi$ represents a wavelet function, $a$ and $b$ denote scaling and translation factors, respectively. It is important to note that although we utilize WT in this work, other approaches like STFT could also be used in conjunction to possibly improve the performance along with segmentation~\cite{sadri2017information}.  

\begin{figure}[!t]
\centering
\includegraphics[width=7.8cm]{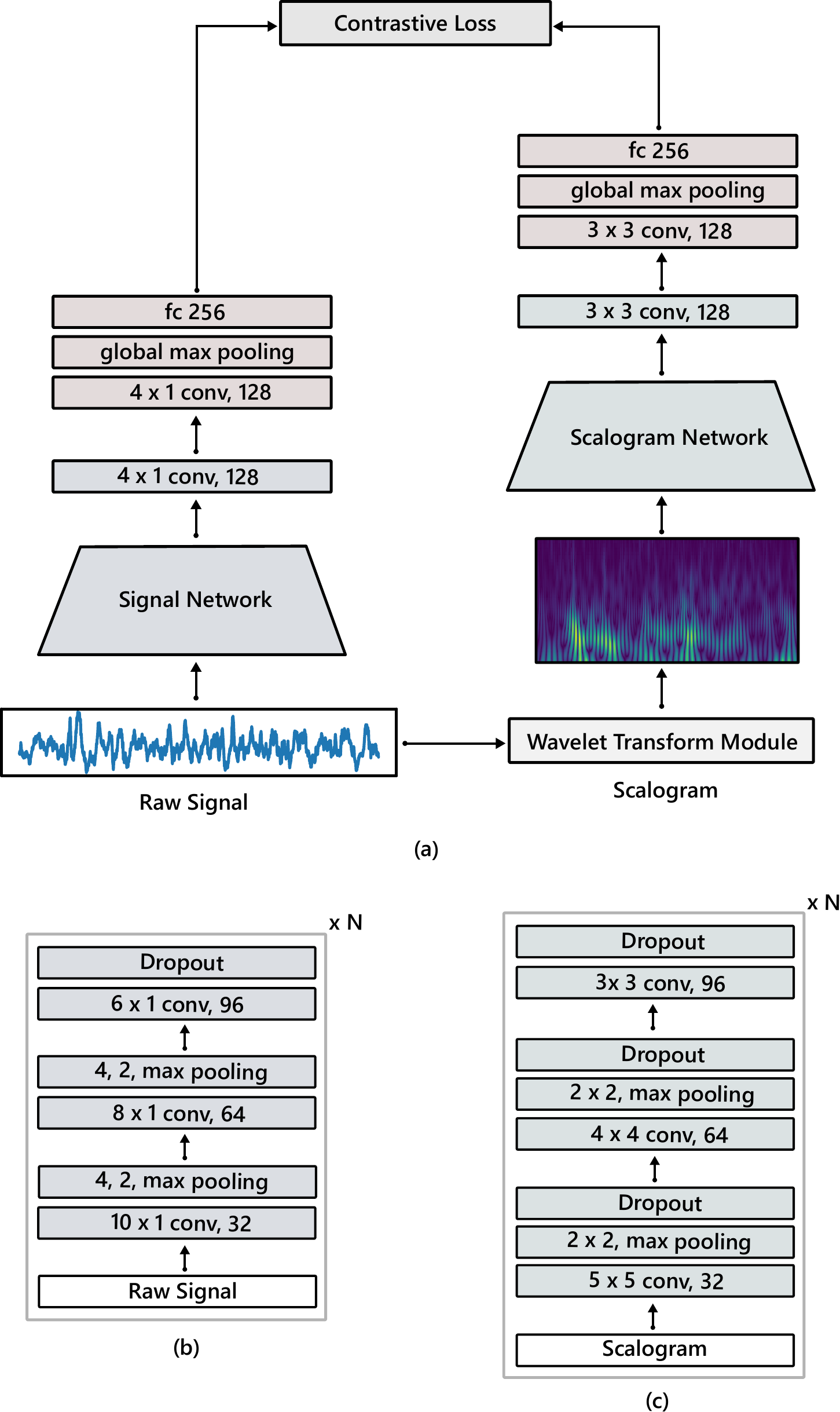}
\caption{\small{Scalogram contrastive network. We design a dual-stream architecture to learn from the raw input signal and its complementary view i.e. a scalogram. We map the original signal fragments into another domain and train the network to recognize which pairs belong together. Within this work, we use a wavelet transform. The high-level overview of the method is illustrated in (a) where signal and scalogram networks are also multi-stream networks with a distinct stream for each input modality. The architecture of these modality-specific signal and scalogram networks is shown in (b) and (c), respectively.}}
\label{fig:wcn}
\end{figure}

\subsection{Federated Learning}
Autonomous vehicles, wearables, smartphones, and IoT sensors are examples of modern distributed devices producing a wealth of data every second. This massive amount of data offers an excellent opportunity for learning models to solve a diverse range of tasks. The applications of interest include customized fitness plans, personalized language models, and contextual awareness for driving automation. The growing computational power of edge devices allows us to leave the data decentralized and push the network computation to the client, which is also ideal from a privacy aspect. The expanding area of federated learning~\cite{44822, li2019federated, kairouz2019advances} explores developing methods to achieve the goal of learning from highly distributed and heterogeneous data through aggregating locally trained models on remote devices, such as smartphones and wearables. In this case, the intention is to minimize the following objective~\cite{li2019federated}:
\begin{align}
   \label{eq:fl}
   \min_{\theta} \mathcal{F}_{\theta}, \ \text{where} \ \ \mathcal{F}_{\theta} := \sum_{c}^{\mathcal{C}} \frac{m_c}{m} \mathcal{F}_{\theta_{c}}.
\end{align}
\noindent Here, $\mathcal{C}$ represents the number of participating client devices in a training round, $m_c$ is the total number of instances available for client $c$ with $m = \sum_{c} m_c$, and lastly $\theta_{c}$ denotes the weights of a local model. To produce a global model, \texttt{Federated Averaging} algorithm~\cite{44822} is typically used to accumulate client updates after every round of local training $t$ as with Equation~\ref{eq:fl}. 

The research interest in federated learning revolves around improving communication efficiency~\cite{konevcny2016federated}, personalization~\cite{wang2019federated}, fault tolerance~\cite{bonawitz2019towards}, privacy preservation~\cite{bonawitz2017practical} as well as looking into the theoretical underpinning of the federated optimization~\cite{li2019federated}. Specifically, the recent work deals with learning a unified model to solve a single as well as multiple tasks~\cite{smith2017federated}. In addition to improving communication costs, the computational efficiency of federated learning on resource-constrained devices has also been studied~\cite{9051991}. Similarly, the development of frameworks and productionizing of applications built around the idea of decentralized learning are also surging to address various practical problems, such as next-word and emoji prediction~\cite{hard2018federated}, wake word recognition~\cite{leroy2019federated}, query suggestion~\cite{yang2018applied}, and traffic flow forecasting~\cite{9082655}. Deep reinforcement learning has also been investigated in a federated setting for edge caching in IoT to improve the quality of services and dealing with traffic off-loading~\cite{9062302}.  

Nevertheless, the existing techniques make a strong assumption that labeled training data are always accessible, or annotations can be extracted reliably, e.g., via user interaction with smartphone applications. However, for various problems involving sensory data (such as sleep stage scoring and context recognition), obtaining a large number of annotated examples in a real-world setting to train supervised models is prohibitively expensive and not feasible. This limits the applicability of current methods in learning from unlabeled data available from distributed IoT devices. The approach presented here is a step towards exploring self-supervised representation learning in a federated setting from unannotated multi-sensor data at the edge. 

\section{APPROACH}
\label{sec:approach}
Learning multi-sensor representations with deep networks requires a large amount of well-curated data, which is made difficult by the diversity of device types, environmental factors, inter-personal differences, privacy issues, and annotation cost. We propose a self-supervised auxiliary task whose objective at a high level is to contrast or compare raw signals and their corresponding scalograms (which are a visual representation of the wavelet transform) so that a network learns to discriminate between aligned and unaligned scalogram-signal pairs. The rationale of the proposed approach is similar in spirit to cross-view learning in the audio-visual domain~\cite{owens2018audio}. However, it differs in a core way that we obtain aligned and unaligned views\footnote{or in-sync and out-of-sync samples} from the same modality with wavelet transform. In the absence of the semantic labels, our methodology can be leveraged to generate an endless stream of labeled data. Therefore, it can train the network without any human involvement, which is particularly attractive for on-device learning. In subsequent sections, we describe details of the correspondence learning, sample generation, preference of a loss function, and key network architectural properties.    

\subsection{Scalogram-signal Correspondence Learning}
\label{sec:sscl}
The idea behind SSCL is to learn network parameters with a self-supervised objective that determines whether a raw signal and a scalogram correspond (or align) with each other or not. Given a multi-sensor dataset with fixed-length input segments of multiple modalities $\mathcal{D} = \{ x_1, x_2, \ldots, x_\mathcal{M} \}$ of $\mathcal{M}$ instances, we train a multimodal contrastive network to achieve the objective of synchronizing representations of the raw input with their corresponding scalogram. Specifically, a time-series is segmented into a fixed-sized input with a sliding window having a certain overlap between samples. Afterward, the scalogram $s_m$ of a signal $x_m$ can be generated with a specified wavelet transformation $\Psi$~\cite{daubechies1990wavelet}. This procedure results in synchronized pairs for each $x_m$ and $s_m$ of $m$-th instance. These co-occurring pairs of inputs are assigned a class label $y_m = 1$, i.e., representing in-sync examples. Likewise, for generating negative samples $y_m = 0$, for a particular $x_m$, a randomly selected $s_m$ is assigned, which in principle represents that these scalogram-signal pairs do not align with each other. Here, we sample a negative scalogram from the same input modality. However, it can also be selected from a different modality, e.g., for accelerometer, the scalogram of the gyroscope can also be utilized. Importantly, we utilize an equal number of positive and negative instances for training the network. As described earlier, a wavelet transform provides a better multi-resolution analysis of non-stationary signals than  Short-Time Fourier Transform (STFT)~\cite{merry2005wavelet}. Hence, we extract a scalogram, which is an absolute and squared value of a WT operation. It is achieved using a continuous Morlet WT function which is expressed as follows:  
\begin{align}
   \label{eq:mwt}
    \psi(t) = \exp^{\frac{-t^{2}}{2}} \cdot \exp^{-jw_{0}t}
\end{align}
\noindent where $w_{0}$ denotes a central frequency of the mother wavelet.

In the broadest sense, the SSCL task requires a semantic understanding of how time-frequency information presented in a scalogram relates to a raw input signal, thus enabling the model to learn general-purpose embedding with a complementary view on the original input. We give a high-level overview of our approach in Figure~\ref{fig:wcn}. The aim here is to learn a classifier $\mathcal{H}(.)$ that can minimize an empirical loss, so $\mathcal{H}(x_{m}, s_{m}) = y_m$. A natural choice is to cast the specified problem as a binary classification task $p(y | x, s)$ and hence, optimize a cross-entropy loss. Nevertheless, we achieve slightly better convergence through employing a contrastive loss that pulls together embedding of positive pairs and pushes different pairs apart, as it is also shown to be improving generalization in earlier work~\cite{chopra2005learning}:
\begin{equation}
\begin{split}
   \mathcal{L} = \frac{1}{\mathcal{M}} \sum_{m=1}^{\mathcal{M}} (y_m) \vert \vert \mathcal{F_{\mathcal{X}}}(x_m) - \mathcal{F_{\mathcal{S}}}(s_m) \vert \vert_2^2 \ \ + \\ (1 - y_m) \max(\alpha - \vert \vert \mathcal{F_{\mathcal{X}}}(x_m) -  \mathcal{F_{\mathcal{S}}}(s_m) \vert \vert_2 ,0)^2
\end{split}
\end{equation}
\noindent where $\alpha$ is a margin hyperparameter, which is enforced between positive and negative samples, $\mathcal{F_{\mathcal{X}}}$, and $\mathcal{F_{\mathcal{S}}}$ are signal and scalogram networks, respectively. The contrastive loss optimization solves the proposed self-supervised task through the integration of not just different views of the same underlying signal, but it also aligns samples across multiple sensory modalities. This label-free correspondence learning approach results in rich representations that may be invariant to sensor noise, amplitude (or scale) variations, user-specific differences, and other factors.

\begin{figure}[t]
\centering
\includegraphics[width=8.6cm]{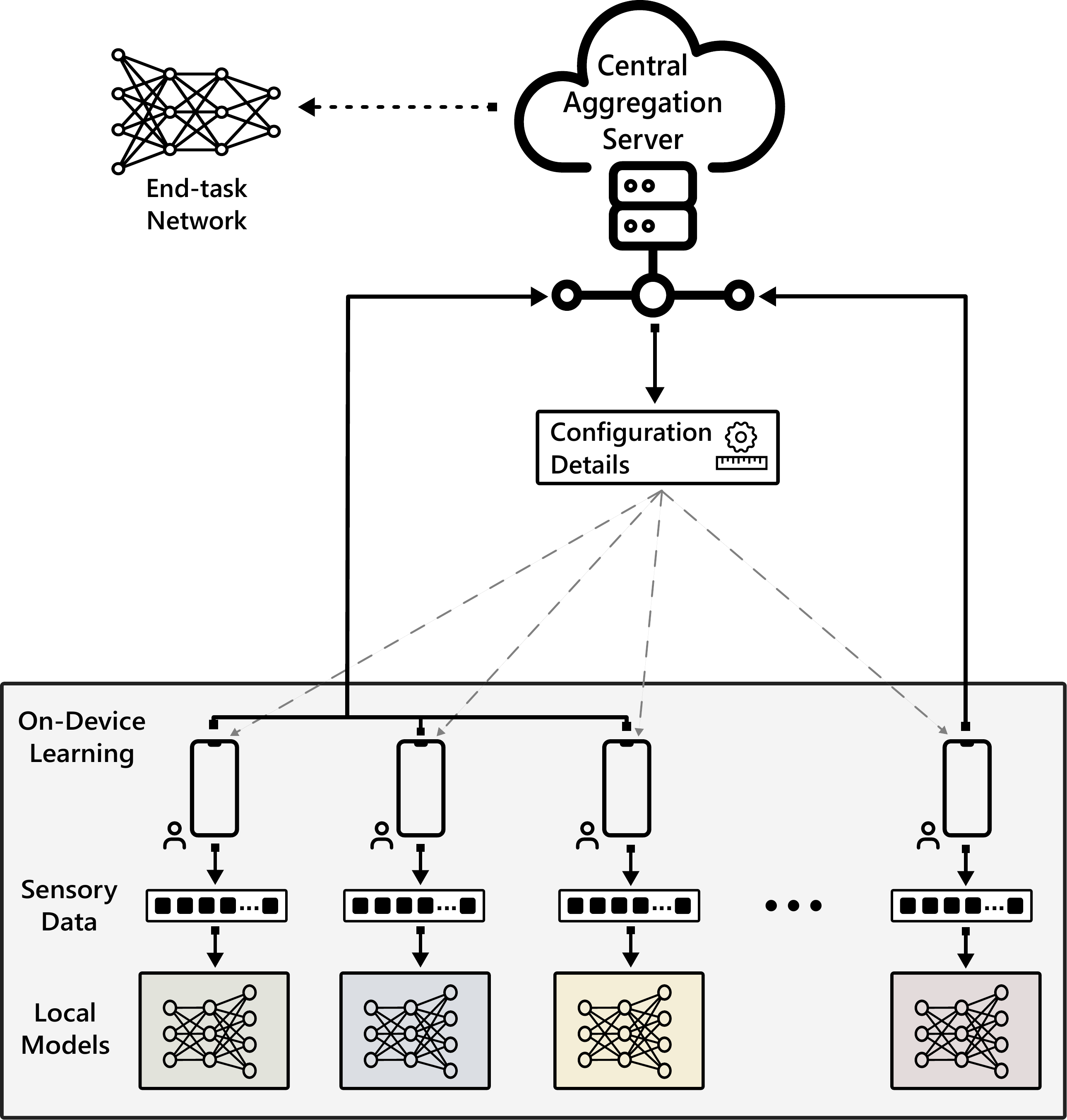}
\caption{\small{Overview of federated learning framework. A central server dispatches a randomly initialized model and other training configuration details to the selected clients' devices, as depicted by dashed gray lines. The clients train local models on their private data and send the models back to the server illustrated with solid black lines. The models are aggregated to produce a unified model that is used for the end-task.}}
\label{fig:overview_fl}
\end{figure}

\subsection{Network Architecture}
To tackle the SSCL task, we design a dual-stream architecture named \textit{scalogram contrastive network}, which is inspired from~\cite{owens2018audio} and it is illustrated in Figure~\ref{fig:wcn}. It is composed of two distinct parts: the scalogram network and the signal network, each extracting features from their respective inputs. As the aim here is to learn representations from multiple sensors, each network consists of modality-specific and fusion layers to learn specialized and joint embedding, respectively. In particular, we utilize the same network architecture for learning on different datasets unless mentioned otherwise. Likewise, only the features from the signal network are used for evaluation, discarding the scalogram network after pre-training. 

The scalogram network consists of three $2$D convolution layers with kernel sizes of $5$, $4$, $3$, and $32$, $64$, $96$ feature maps, respectively. Dropout is applied after every layer and max-pooling after the initial two convolutional layers with a pooling size of $2$. We use the same design for each input modality, followed by the fusion layer consisting of $128$ feature maps with a kernel size of $3$. To learn from raw signals, we use a $1$D convolutional network with the same structure as the scalogram network but with crucial differences in kernel sizes which are $10$, $8$, and $6$ for sensor-specific layers and $4$ in the case of a shared layer with a dropout layer at the end. Moreover, we use additional pre-training related layers for both networks, comprising a convolutional layer with $128$ feature maps and a dense layer with $256$ hidden units. These layers are discarded after the self-supervised learning phase as we hypothesize that they might learn features relevant to the auxiliary task (i.e., SSCL). We use the Mish~\cite{misra2019mish} activation function in all the layers except the last, which has either linear or softmax activation. Finally, the input to our scalogram network are coefficients of the wavelet transform with a size ($h \times w \times c$), each representing height, width, and the number of channels, respectively. The signal network directly processes raw input of size ($w \times c$). 

\subsection{Implementation Details}
For pre-training, we sample the non-corresponding scalogram-signal examples through randomly selecting scalograms from outside the current input batch while keeping the raw input fixed for positives and negatives. We preprocess the signals before computing scalogram or initiating network training as done in the previous works for each considered dataset; further details are provided in Section~\ref{sec:datasets}. We calculate summary statistics for z-normalization from the training set. We use an Adam optimizer with a fixed learning rate of $0.0001$ for pre-training and $0.01$ or $0.02$ in case of learning a linear classifier, which could also be decayed based on performance on the validation set. The network is trained with a batch size of $24$, a dropout rate of $0.1$, and L$2$ regularization rate of $0.0001$. For federated learning simulation, we use the Tensorflow federated learning framework\footnote{https://www.tensorflow.org/federated}. In this case, the networks are trained with a batch size of $12$ for $5$ local epochs using data of $n$ randomly selected users (typically $10$) at each training round with $30-50$ rounds in total, depending on the dataset size. Specifically, in our experiments, we randomly divide the training set into multiple subsets (representing each client) that are used to train the models in a federated setting. We opt for this strategy due to fewer users in existing datasets. The availability of bigger datasets with a larger pool of users could be useful to evaluate self-supervised methods in the future. A high-level overview of federated learning is illustrated in Figure~\ref{fig:overview_fl}. 

\begin{table}[t]
\centering
\small
\def\arraystretch{1.5}
\caption{\small{Summary of datasets.}}
\begin{tabular}{c|c|c|c}
\textbf{Task} & \textbf{Dataset} & \textbf{\#Users} & \textbf{\#Outputs} \\ \hline
Sleep Stage Scoring & Sleep-EDF & 20 & 5 \\ \hline
\multirow{2}{*}{Activity Recognition} & HHAR & 9 & 6 \\ 
 & MobiAct & 61 & 11 \\ \hline
Device-Free Sensing & WiFi-CSI & 6 & 7 \\ \hline
Stress Detection & WESAD & 15 & 3 \\ 
\end{tabular}
\label{tab:data_details}
\end{table}

\section{EXPERIMENTS}
We evaluate the effectiveness of our approach in multiple ways with several publicly available datasets from different domains. First, we probe the quality of representations with a linear classifier trained on-top of a frozen feature extractor in both central and federated learning settings. Second, we examine whether scalogram-signal correspondence learning could be used to improve the recognition rate in the low-data regime. Finally, we determine the transferability of features on related datasets, followed by an evaluation with cross-validation to determine robustness against subject variations. 

\subsection{Datasets and Preprocessing}
\label{sec:datasets}
We experimented with learning models on $5$ datasets from the following application areas: sleep stage scoring, human activity recognition, WiFi sensing, and physiological stress detection. 

The electroencephalogram (EEG) and electrooculography (EOG) signals are used from the PhysioNet Sleep-EDF dataset~\cite{kemp2000analysis,goldberger2000physiobank} for classifying sleep into five stages (i.e., Wake, N$1$, N$2$, N$3$, and  Rapid Eye Movement). We preprocess these signals, which are recorded at $100$Hz, as done in earlier work~\cite{supratak2017deepsleepnet} and utilize 30-second epochs (segments). For activity classification with smartphones, accelerometer, and gyroscope signals from HHAR~\cite{stisen2015smart} and MobiAct~\cite{vavoulas2016mobiact} datasets are used, which have $6$ and $11$ output classes, respectively. We segment the raw signals through a sliding window into a segment size of $400$ samples with a $50\%$ overlap. For device-free sensing of daily activities, we use the WiFi channel state information data~\cite{yousefi2017survey} and follow identical preprocessing steps with~\cite{yousefi2017survey}. Notably, the signals are resampled from $1$kHz to $500$Hz through uniform temporal downsampling with a rate of $2$ for each of the $90$ channels (i.e., $30$ sub-carriers per antenna) to classify them into $7$ classes. The WESAD dataset~\cite{schmidt2018introducing} is used for the detection of stress, normal, and amusement physiological states. Here, we use blood volume pulse, electrodermal activity, and temperature signals collected from a wrist wearable device at $64$Hz, $4$Hz, and $4$Hz, respectively. Following~\cite{schmidt2018introducing}, we extract $30$-seconds segments and independently normalize each subject's data before the model development phase. 

\begin{table*}[t]
\small
\centering
\def\arraystretch{1.5}
\caption{\small{Performance evaluation of self-supervised representations learned in a standard central setting with a linear classifier.}}
\label{tab:std_results}
\begin{tabular}{c|cc|cc|cc|cc|cc}
 & \multicolumn{2}{|c|}{\textbf{Sleep-EDF}} & \multicolumn{2}{c|}{\textbf{HHAR}} & \multicolumn{2}{c|}{\textbf{MobiAct}} & \multicolumn{2}{c|}{\textbf{WiFi-CSI}} & \multicolumn{2}{c}{\textbf{WESAD}} \\ \hline
 & F-score & Kappa & F-score & Kappa & F-score & Kappa & F-score & Kappa & F-score & Kappa \\ \hline
Random Init. & 0.67 & 0.54 & 0.64 & 0.58 & 0.65 & 0.63 & 0.36 & 0.24 & 0.73 & 0.58 \\ \hline
Supervised & 0.82 & 0.76 & 0.73 & 0.69 & 0.95 & 0.93 & 0.96 & 0.95 & 0.85 & 0.75 \\ \hline
Autoencoder & 0.75 & 0.66 & 0.69 & 0.63 & 0.80 & 0.78 & 0.84 & 0.81 & 0.83 & 0.72 \\
\hline
SCN & 0.78 & 0.70 & 0.82 & 0.79 & 0.91 & 0.88 & 0.84 & 0.81 & 0.84 & 0.73 \\
\end{tabular}
\end{table*}

\begin{table*}[t]
\small
\centering
\def\arraystretch{1.5}
\caption{\small{Assessing performance in a federated learning setting to determine SCN's ability to learn representations from distributed data. The entries marked with FC (federated classifier) denotes metrics when both representations and classifier are learned in a federated context.}}
\label{tab:fed_results}
\begin{tabular}{c|cc|cc|cc|cc|cc}
 & \multicolumn{2}{|c|}{\textbf{Sleep-EDF}} & \multicolumn{2}{c|}{\textbf{HHAR}} & \multicolumn{2}{c|}{\textbf{MobiAct}} & \multicolumn{2}{c|}{\textbf{WiFi-CSI}} & \multicolumn{2}{c}{\textbf{WESAD}} \\ \hline
 & F-score & Kappa & F-score & Kappa & F-score & Kappa & F-score & Kappa & F-score & Kappa \\ \hline
Supervised & 0.82 & 0.76 & 0.77 & 0.73 & 0.94 & 0.92 & 0.92 & 0.90 & 0.85 & 0.75 \\ \hline
Autoencoder & 0.76 & 0.68 & 0.71 & 0.66 & 0.86 & 0.83 & 0.85 & 0.81 & 0.82 & 0.70 \\\hline
SCN & 0.78 & 0.70 & 0.80 & 0.77 & 0.90 & 0.88 & 0.85 & 0.82 & 0.83 & 0.73 \\ \hline
Autoencoder (FC) &  0.68 & 0.56 &  0.51 & 0.44 & 0.54 & 0.47 & 0.67 & 0.60 & 0.80 & 0.67 \\ \hline
SCN (FC) &  0.77 & 0.69 & 0.80 & 0.76 & 0.82 & 0.79 & 0.69 & 0.63 & 0.82 & 0.70 \\
\end{tabular}
\end{table*}

In all the cases, we use a random $70\%-30\%$ split of the dataset (based on users such that there is no overlap in terms of users' data) for training and evaluation, respectively. We also pick a $20\%$ subset from training split as a validation set for hyperparameter tuning and model selection. Moreover, we also evaluate the performance of our approach with cross-validation based on user split, i.e., leave-one-user-out. Table~\ref{tab:data_details} summarizes the key characteristics of the datasets used in our evaluation. 

\begin{figure}[htbp]
\centering
\includegraphics[width=8.2cm]{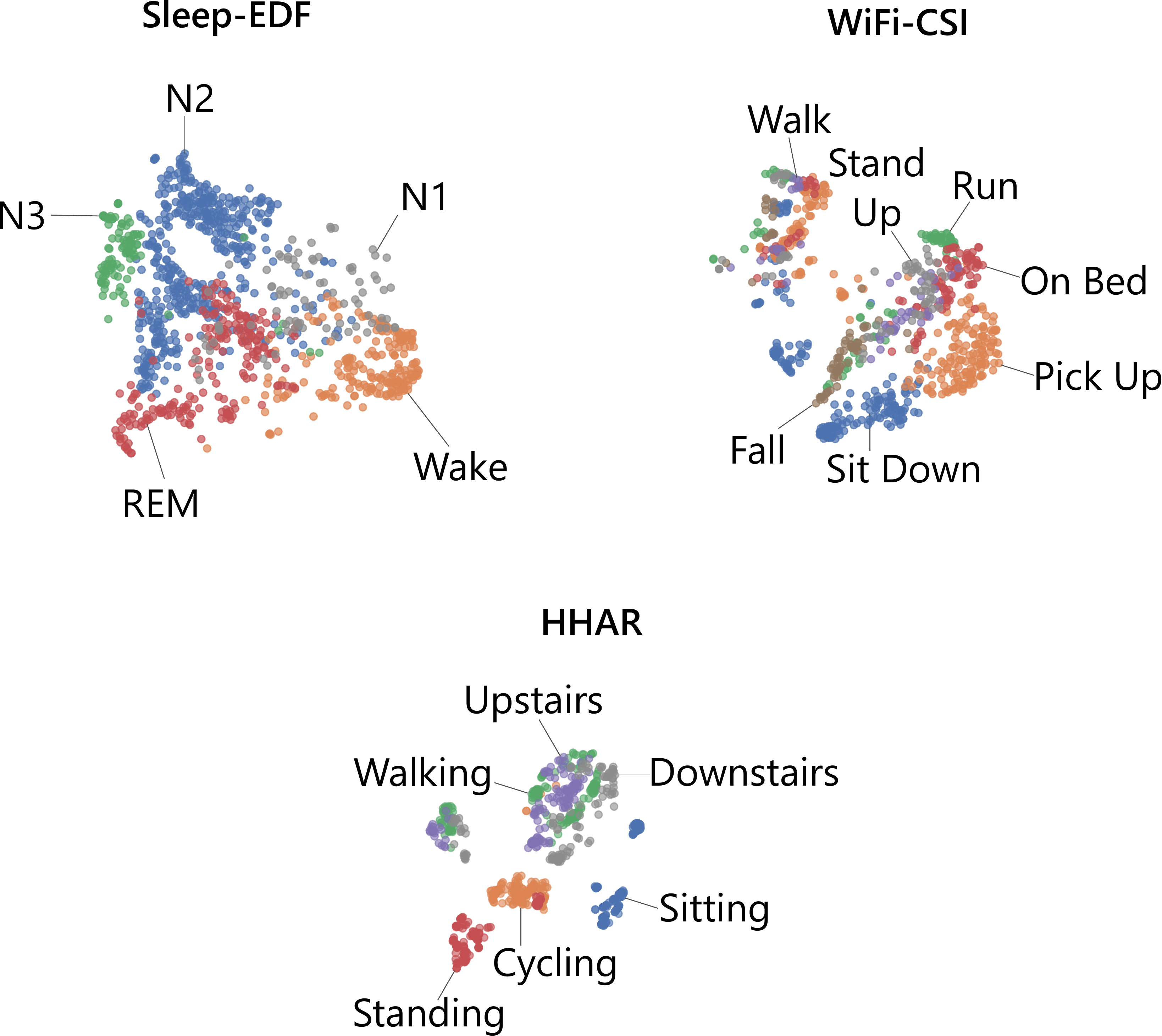} 
\caption{\small{t-SNE embedding learned with scalogram contrastive network on a random subset of test subjects. Note, t-SNE does not utilize class labels, the colors are added during post-hoc analysis for better interpretability.}}
\label{fig:tsne_results}
\end{figure}

\subsection{Quality assessment of the learned features with separability analysis}
In Table~\ref{tab:std_results} and Table~\ref{tab:fed_results}, we provide our key evaluation results in central and federated learning settings. First, we compare the performance of our approach with a) a supervised network trained end-to-end, b) an autoencoder, and c) a randomly initialized network in a central setting, i.e. when the entire data are available for learning on a server. We measure the quality of learned representations through a linear classifier trained on-top of the frozen feature extractor, which is a standard evaluation protocol used in earlier work. In the federated setting (Table~\ref{tab:fed_results}, the supervised network is learned for each user, and the weights are aggregated to create a unified model. For an autoencoder and SCN, the pre-training is performed in a federated setting to learn representations, and a classifier is trained in a standard way i.e., as if the data of end-task are available on the server. In addition, we also assess the performance when unsupervised networks are kept frozen, and classifier is also learned in a federated setting. In Table~\ref{tab:std_results} these entries are represented with FC, which is an abbreviation of a federated classifier. 

In particular, we highlight that for federated learning, we utilize random partitioning of the training sets as in~\cite{44822} to tackle the low number of users in the considered (existing) datasets. This choice might result in a decentralized IID (i.e., independent and identically distributed) dataset that could be unbalanced but does not suffer from extreme heterogeneity in terms of training instances per client as generally, the case is for non-IID data that typically varies heavily based on the users' demographics, device usage, and other factors. However, we would again emphasize that our self-supervised technique does not depend on the user-generated labels for learning representations and could be easily applied to large-scale datasets. However, as the end-task labels are required to evaluate the quality of learned features, the unavailability of massive multi-sensor labeled data is a critical limiting factor towards realizing the goal of assessment in the non-IID setting. We leave the evaluation of self-supervised features on a large pool of users with a greater variety of devices as future work. 

\begin{figure}[!htbp]
\centering
\includegraphics[width=7.8cm]{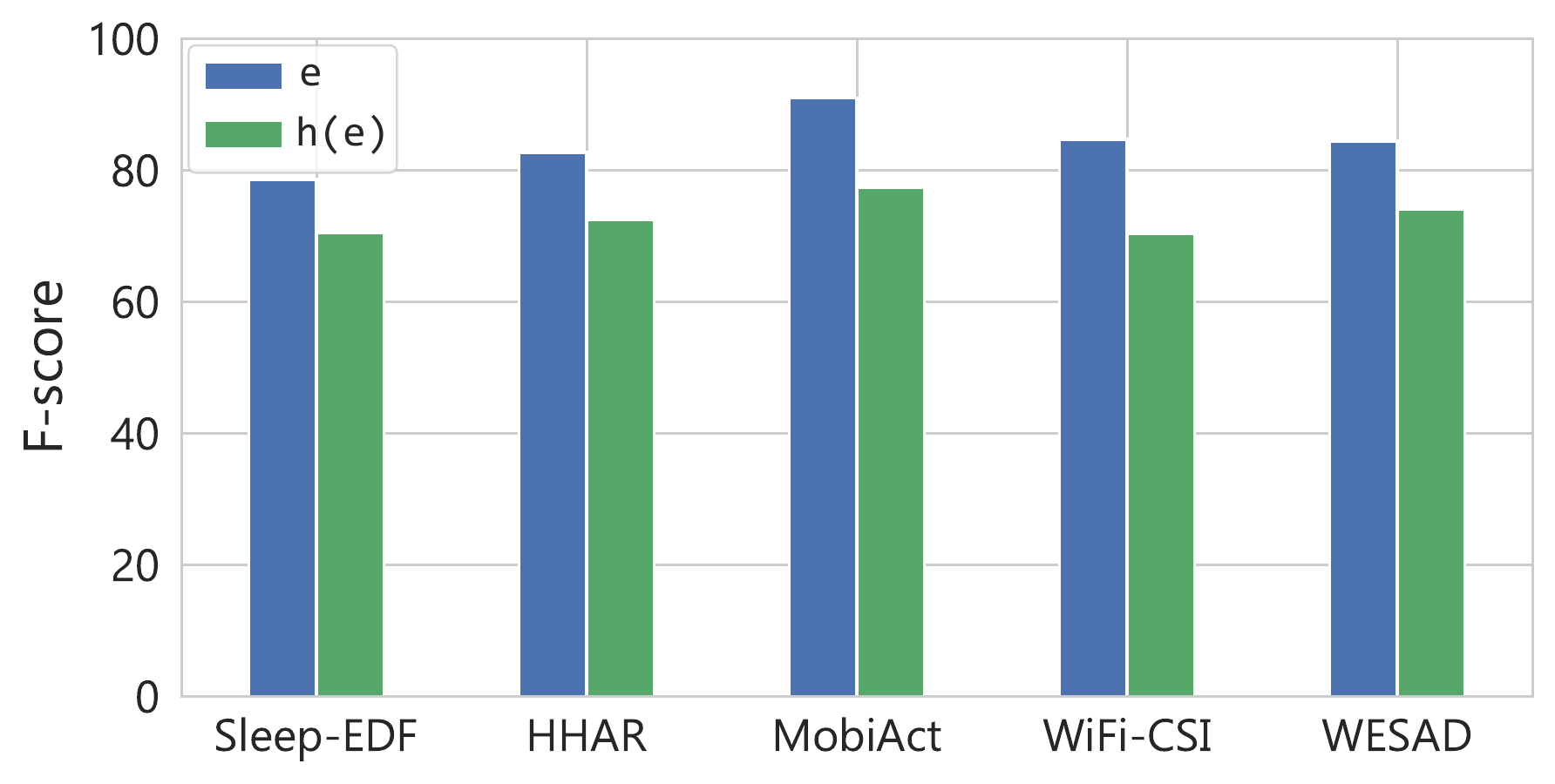} 
\caption{\small{Performance comparison of linear classifiers trained on-top of representations from encoder $(e)$ and penultimate layer's projection $h(e)$ of SCN denoted with \texttt{fc 256} in Figure~\ref{fig:wcn}.}}
\label{fig:e_h_results}
\end{figure}

\begin{figure*}[t]
\centering
\subfloat{\includegraphics[width=4.7cm]{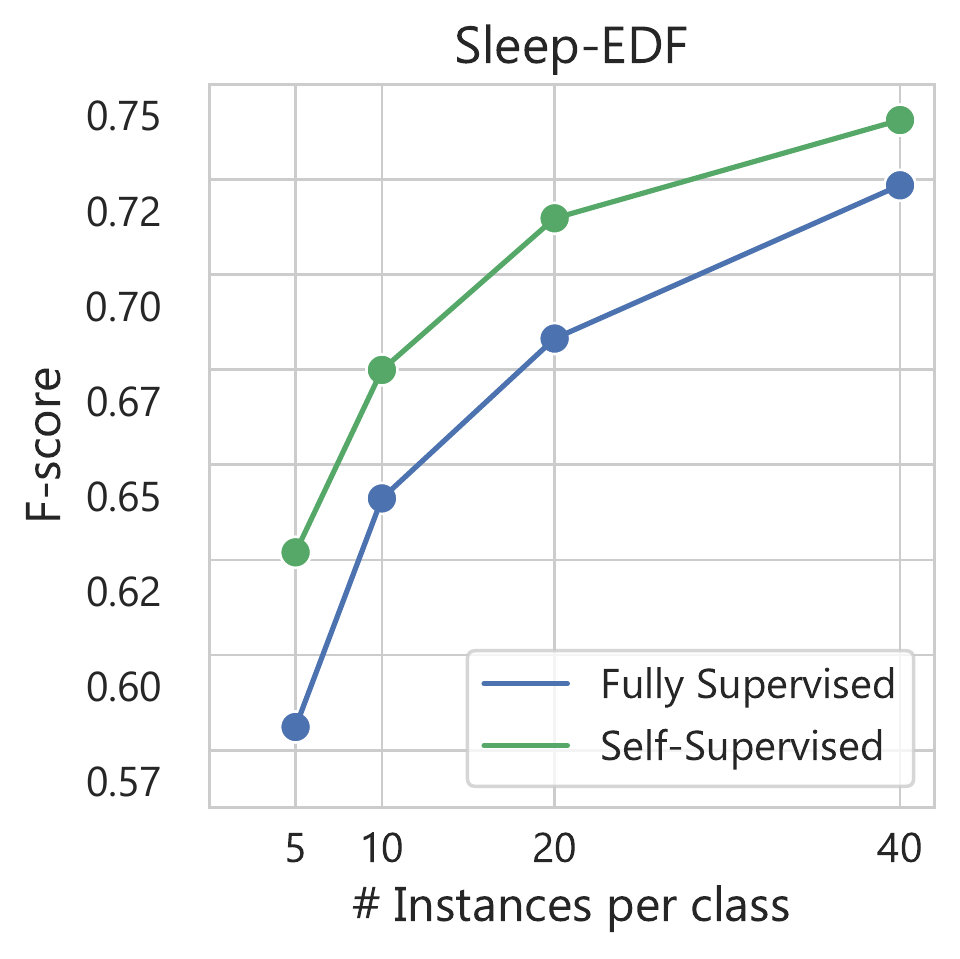}}  
\subfloat{\includegraphics[width=4.7cm]{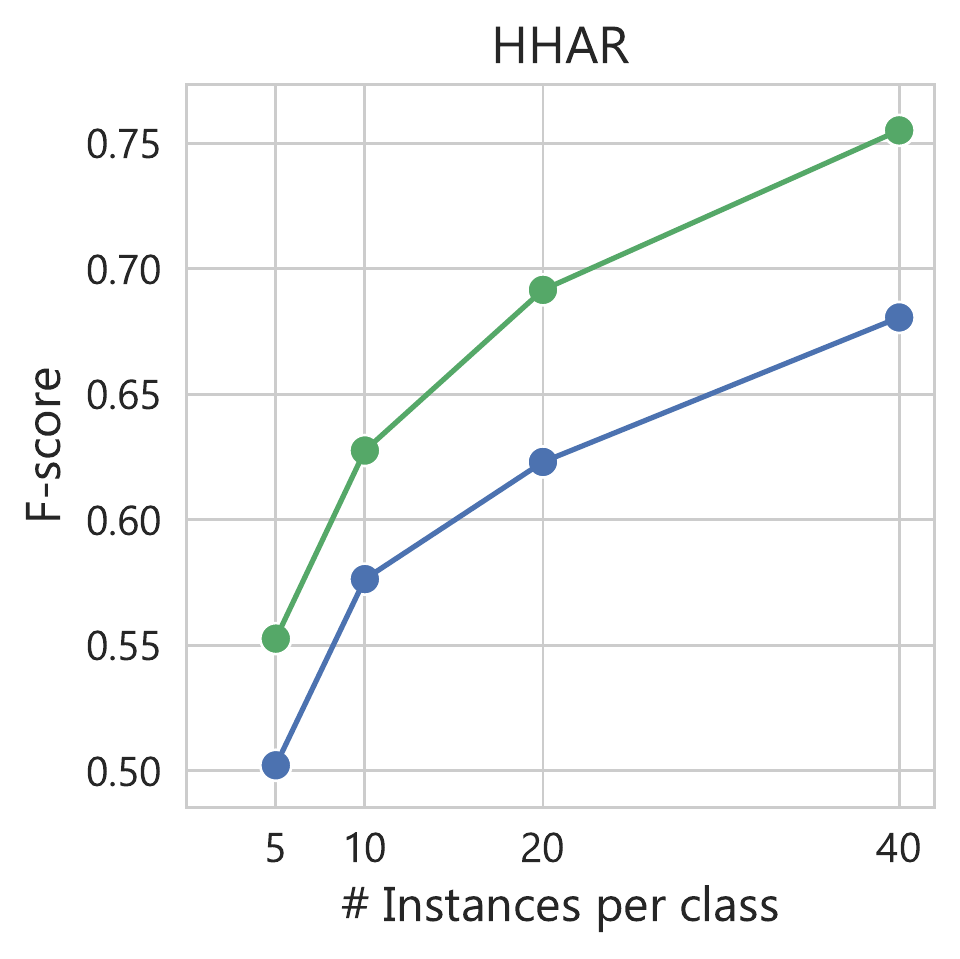}} 
\subfloat{\includegraphics[width=4.7cm]{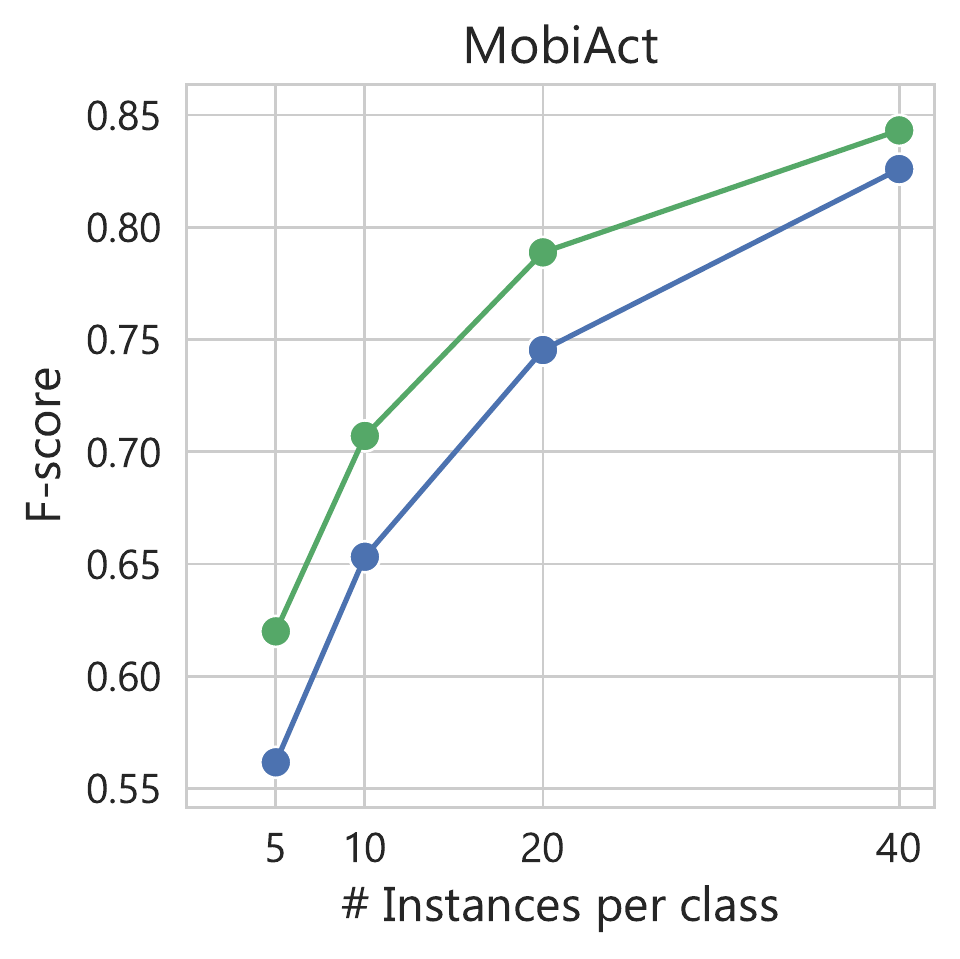}} \\
\subfloat{\includegraphics[width=4.7cm]{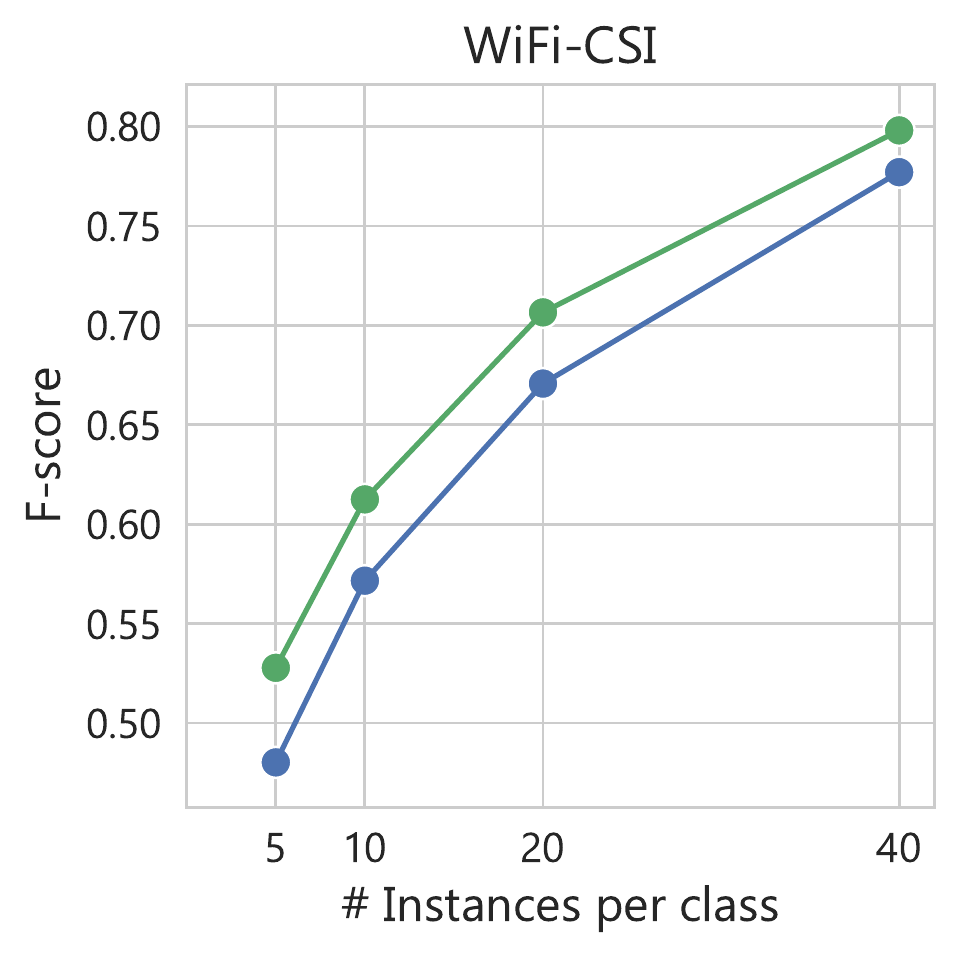}} 
\subfloat{\includegraphics[width=4.7cm]{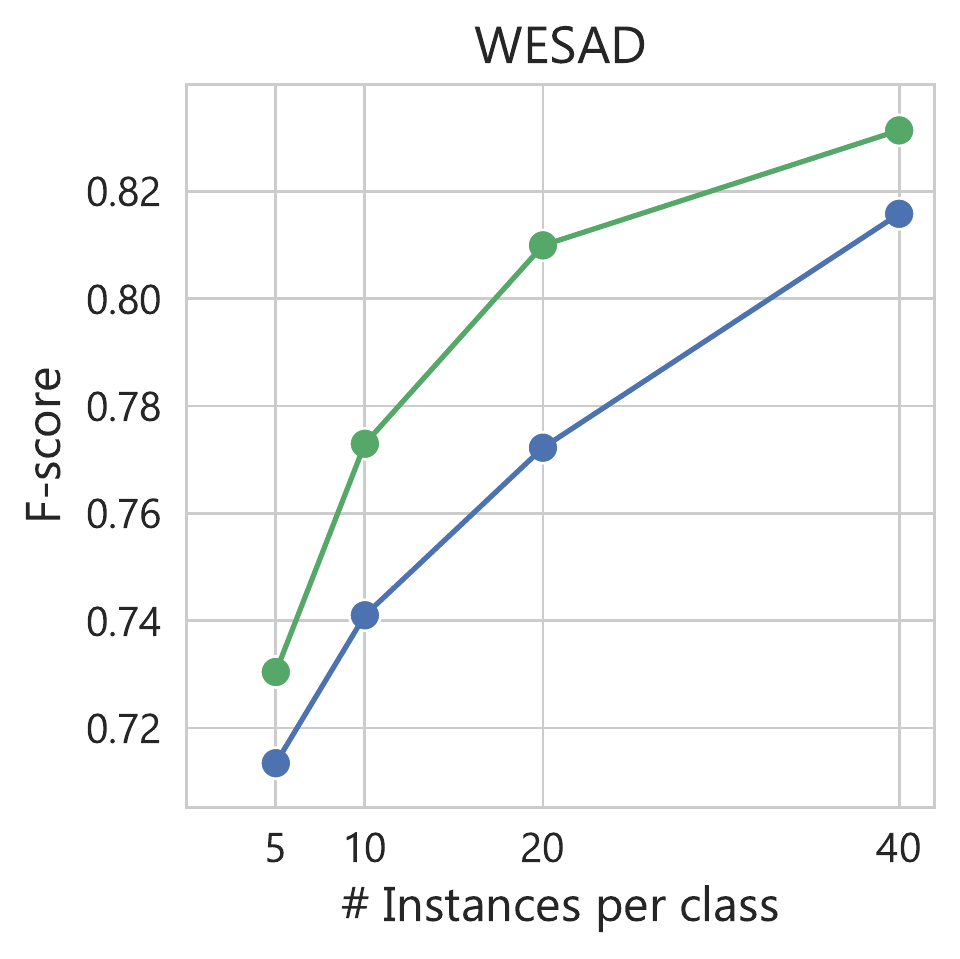}} 
\caption{\small{Effectiveness of self-supervised learning in a low-data regime. The SCN is pre-trained on unlabeled data and fine-tuned end-to-end with few-labeled data points (i.e, $5, 10, 20, \text{and } 40$ instances per class). On all the evaluated datasets, we notice a significant performance improvement over a supervised baseline network, which is trained only with labeled inputs.}}
\label{fig:ld_results}
\end{figure*}

On the evaluated datasets, we observe that the classifiers learned on-top of a fixed randomly initialized network achieve F-score above $60\%$ in most cases. It highlights the representational capacity of our architecture design that, without seeing any samples, the encoder can provide reasonable embedding for a linear classifier. Notably, the SCN surpasses pre-training results with the autoencoder and on HHAR achieves better F-score ($82.7$) than a supervised baseline ($73.0$). Particularly, we notice that the results obtained in a federated setting are close to those achieved with learning end-to-end models in a central setting, which hints towards the robustness of our approach in a federated environment. Similarly, when a linear classifier is also trained in a federated setting, the performance of SCN is mainly consistent with the centralized classifier, which is not the case for an autoencoder. Moreover, in Figure~\ref{fig:tsne_results}, we provide the t-SNE embedding of SCN on $1000$ randomly selected instances from a test set of Sleep-EDF, WiFi-CSI, and HHAR. The distinct clusters of data points can be seen that are discovered entirely in an unsupervised manner. This further highlights the ability of SCN to learn meaningful representations. 

In Figure~\ref{fig:e_h_results}, we compare the performance of downstream task classifiers trained on embedding from two different parts of the network. The representations from the encoder $e$ and the features from the penultimate layer of SCN $h(e)$ are used for this purpose. It can be seen that the classifier trained on the output of $e$ performs significantly better than the one learned using the last layer's features. We think it could be because that layers at the end might learn auxiliary task-specific features that are not useful enough for the end-task.

\subsection{Improving generalization in low-data regime and transfer as evaluation}
We explore the effectiveness of the proposed technique for improving performance with few-labeled examples. We pre-train a scalogram-contrastive network with the entire unlabeled data and use the model as initialization for learning a downstream task. We compare the performance with a standard supervised network trained only with certain labeled instances. Specifically, we use $5$, $10$, $20$, and $40$ labeled instances per class to learn the end-task model. Figure~\ref{fig:ld_results} and Table~\ref{tab:ld_results} show an average F-score of $100$ independent repetitions where different examples are sampled to train the network at each run. In all the cases, the results obtained with utilizing a self-supervised network are better than the baseline, even when limited labeled data are available. This highlights that the SCN efficiently harnesses unlabeled data to learn generalized features.

Similarly, the self-supervised networks are also evaluated in terms of their usefulness in a transfer learning setting. Generally, this is achieved by treating a pre-trained model as a fixed feature extractor, and a linear model is trained on top of it using a different dataset. Here, we assess the performance on activity recognition tasks with HHAR and MobitAct datasets. Table~\ref{tab:tf_results} provides these results and compares with the supervised network, transfer from supervised (Sup.), and SCN trained on the same source instances. In both cases, we see that the recognition improves relatively if the transferred embedding is from SCN compared to a supervised network. Finally, we also assess the performance of SCN when few-labeled instances are available for fine-tuning, but different unlabeled data are available for pre-training, as shown in Table~\ref{tab:tf_sm_results}. Similar to earlier semi-supervised evaluation, we fine-tune a pre-trained network end-to-end with $5, 10, 20, \text{and } 40$ examples of each class from the target dataset. We notice a $2\%-3\%$ improvement in F-score over the supervised network when an SCN encoder is utilized.

\begin{table*}[!htbp]
\centering
\def\arraystretch{1.5}
\caption{Generalization improvement in semi-supervised setting with self-supervised pre-training.}
\label{tab:ld_results}
\begin{tabular}{c|cc|cc|cc|cc|cc}
 & \multicolumn{2}{|c}{\textbf{Sleep-EDF}} & \multicolumn{2}{|c}{\textbf{HHAR}} & \multicolumn{2}{|c}{\textbf{MobiAct}} & \multicolumn{2}{|c}{\textbf{WiFi-CSI}} & \multicolumn{2}{|c}{\textbf{WESAD}} \\ \hline
 & Supervised & SCN & Supervised & SCN & Supervised & SCN & Supervised & SCN & Supervised & SCN \\ \hline
5 & 0.58$\pm$0.05 & 0.62$\pm$0.05 & 0.50$\pm$0.07 & 0.55$\pm$0.06 & 0.56$\pm$0.06 & 0.61$\pm$0.07 & 0.48$\pm$0.03 & 0.52$\pm$0.03 & 0.71$\pm$0.06 & 0.73$\pm$0.06 \\ \hline
10 & 0.64$\pm$0.03 & 0.67$\pm$0.04 & 0.57$\pm$0.06 & 0.62$\pm$0.05 & 0.65$\pm$0.05 & 0.70$\pm$0.05 & 0.57$\pm$0.02 & 0.61$\pm$0.02 & 0.74$\pm$0.03 & 0.77$\pm$0.03 \\ \hline
20 & 0.68$\pm$0.05 & 0.71$\pm$0.02 & 0.62$\pm$0.05 & 0.69$\pm$0.04 & 0.74$\pm$0.04 & 0.78$\pm$0.04 & 0.67$\pm$0.02 & 0.70$\pm$0.02 & 0.77$\pm$0.03 & 0.80$\pm$0.03 \\ \hline
40 & 0.72$\pm$0.03 & 0.74$\pm$0.02 & 0.68$\pm$0.04 & 0.75$\pm$0.04 & 0.82$\pm$0.02 & 0.84$\pm$0.02 & 0.77$\pm$0.02 & 0.79$\pm$0.02 & 0.81$\pm$0.02 & 0.83$\pm$0.02
\end{tabular}
\end{table*}

\begin{table}[htbp]
\centering
\def\arraystretch{1.5}
\caption{\small{Evaluation of self-supervised representation in a standard transfer learning setting.}}
\label{tab:tf_results}
\small
\begin{tabular}{c|cc|cc}
 & \multicolumn{2}{c}{\textbf{HHAR $\rightarrow$ MobiAct}} & \multicolumn{2}{|c}{\textbf{MobiAct $\rightarrow$ HHAR}} \\ \hline
 & F-score & Kappa & F-score & Kappa \\ \hline
\begin{tabular}[c]{@{}c@{}}Supervised \end{tabular} &  0.95 & 0.93 & 0.73 & 0.69 \\ \hline
\begin{tabular}[c]{@{}c@{}}Source (SCN) \end{tabular} & 0.91 & 0.88 & 0.82 & 0.79 \\ \hline
\begin{tabular}[c]{@{}c@{}}Transfer (Sup.) \end{tabular} & 0.86 & 0.83 & 0.62 & 0.54 \\ \hline
\begin{tabular}[c]{@{}c@{}}Transfer (SCN) \end{tabular} & 0.87 & 0.84 & 0.75 & 0.71 \\
\end{tabular}
\end{table}

\begin{table}[htbp]
\centering
\def\arraystretch{1.5}
\caption{\small{Fine-tuning transferred model with few-labeled data to improve recognition rate. We report weighted F-score averaged over $100$ independent runs. T denotes a transfer learning.}}
\small
\begin{tabular}{c|cc|cc}
& \multicolumn{2}{c|}{\textbf{HHAR $\rightarrow$ MobiAct}} & \multicolumn{2}{c}{\textbf{MobiAct $\rightarrow$ HHAR}} \\ \hline
 & Supervised & SCN (T) & Supervised & SCN (T) \\ \hline
5 & 0.50 & 0.51 & 0.56 & 0.59 \\ \hline
10 & 0.56 & 0.59 & 0.65 & 0.69 \\ \hline
20 & 0.62 & 0.65 & 0.74 & 0.75 \\ \hline
40 & 0.68 & 0.70 & 0.82 & 0.82
\end{tabular}
\label{tab:tf_sm_results}
\end{table}

\begin{table*}[htbp]
\centering
\def\arraystretch{1.5}
\caption{\small{Comparison of self-supervised representations to a fully-supervised network and pre-training with autoencoder using cross-validation.}}
\label{tab:cv_results}
\small
\begin{tabular}{c|cc|cc|cc}
 & \multicolumn{2}{c|}{\textbf{Supervised}} & \multicolumn{2}{c|}{\textbf{Autoencoder}} & \multicolumn{2}{c}{\textbf{SCN}} \\ \hline
 & F-score & Kappa & F-score & Kappa & F-score & Kappa \\  \hline
Sleep-EDF & 0.83$\pm$0.05 & 0.77$\pm$0.06 & 0.73$\pm$0.08 & 0.65$\pm$0.10 & 0.82$\pm$0.03 & 0.83$\pm$0.03 \\ \hline
HHAR & 0.82$\pm$0.12 & 0.80$\pm$0.13 & 0.62$\pm$0.13 & 0.59$\pm$0.15 & 0.78$\pm$0.11 & 0.76$\pm$0.12 \\ \hline
MobiAct & 0.94$\pm$0.02 & 0.92$\pm$0.03 & 0.79$\pm$0.04 & 0.75$\pm$0.06 & 0.90$\pm$0.02 & 0.87$\pm$0.03 \\ \hline
WiFi-CSI & 0.97$\pm$0.0 & 0.97$\pm$0.0 & 0.85$\pm$0.01 & 0.82$\pm$0.02 & 0.85$\pm$0.01 & 0.82$\pm$0.01 \\ \hline
WESAD & 0.76$\pm$0.11 & 0.63$\pm$0.17 & 0.71$\pm$0.14 & 0.56$\pm$0.25 & 0.75$\pm$0.13 & 0.63$\pm$0.19 \\ 
\end{tabular}
\end{table*}

\subsection{Network robustness against subject variation with cross-validation}
To determine the robustness of network pre-training with the proposed approach against subject variation, we perform cross-validation (CV) based on user split. For Sleep-EDF, HHAR, and WESAD leave-one-subject-out CV is employed, whereas for MobiAct and WiFi-CSI, a $10$-fold stratified CV is used due to a large number of users in the former and unavailability of subject ID's in the latter. We follow the same evaluation strategy as earlier, i.e., training a linear classifier to assess the quality of representations compared to the fully-supervised model and an autoencoder. Table~\ref{tab:cv_results} summarizes mean and standard deviation of metrics averaged over folds. Overall, we notice that SCN is stable despite the changes of subject data in a training set and achieves significantly better results than an autoencoder. Notably, on Sleep-EDF, our methods achieve a mean kappa score of $0.83$ as compared to $0.77$ of a supervised network and $0.76$ as reported in~\cite{supratak2017deepsleepnet}. Likewise, our self-supervised technique performs better than the hand-designed features from wrist physiological signals on WESAD by achieving an F-score of $75.7 \pm 0.13$ as compared to $66.33 \pm 0.36$~\cite{schmidt2018introducing}. Furthermore, we would like to highlight that a direct comparison of existing approaches on other datasets used in our study is not feasible due to the differences in reported metrics and used sensing modalities. Nevertheless, our results with cross-validation further indicate that self-supervised learning can be effectively utilized for sensor modeling tasks on a large scale and can be combined with active learning methods \cite{liono2019improving}. 

\section{CONCLUSIONS}
In this paper, we propose a self-supervised method for learning representations from unlabeled multi-sensor input data, which is typical in the IoT setting. Our method utilizes wavelet transform to generate a complementary view of the input (i.e., a scalogram) to define an auxiliary task of scalogram-signal correspondence. This procedure is specifically designed to work in federated learning setting to allow training networks with widely distributed and unannotated data as the labels can be readily extracted from the data without human-in-the-loop. We show the efficacy of the developed technique on several publicly available datasets involving diverse sensory streams, such as electroencephalogram, blood volume pulse, and IMUs. Particularly, we evaluate the quality of learned features with a linear classifier on an end-task and compare the performance with a fully-supervised network and pre-training with an autoencoder in both federated and central settings. Furthermore, we demonstrate an improved generalization in the low-data regime with self-supervision, i.e., when few labeled instances are used for fine-tuning network on the desired end-task. Our generic self-supervised approach can be used efficiently to learn general-purpose deep feature extractors entirely on-device without the need to transmit the actual data to the server. In future work, we plan to combine self-supervision with architecture search on larger datasets and evaluate our method in a non-IID setting for federated learning. Another avenue of future research is to explore the effectiveness of self-supervised pre-training for adversarial robustness in a federated setting.

\bibliographystyle{IEEEtran}
\bibliography{main}
\end{document}